\documentclass{article}
\usepackage{ijcai26}

\usepackage{times}
\usepackage{soul}
\usepackage{url}
\usepackage[hidelinks]{hyperref}
\usepackage[utf8]{inputenc}
\usepackage[small]{caption}
\usepackage{graphicx}
\usepackage{amsmath}
\usepackage{amsfonts}
\usepackage{amsthm}
\usepackage{booktabs}
\usepackage{algorithm}
\usepackage{algorithmic}
\usepackage[switch]{lineno}
 \usepackage{multirow}

\title{Doc2AHP: Inferring Structured Multi-Criteria Decision Models via Semantic Trees with LLMs}

\author{
Hongjia Wu$^1$
\and
Shuai Zhou$^2$\and
Hongxin Zhang$^3$\And
Wei Chen$^4$\\
\affiliations
$^1$State Key Laboratory of CAD\&CG, Zhejiang University, Hangzhou\\
\emails
\{hongjiawu, zhoushua1, zhx, chenvis\}@zju.edu.cn
}

\begin{document}

\maketitle

\begin{abstract}
    While Large Language Models (LLMs) demonstrate remarkable proficiency in semantic understanding, they often struggle to ensure structural consistency and reasoning reliability in complex decision-making tasks that demand rigorous logic. Although classical decision theories, such as the Analytic Hierarchy Process (AHP), offer systematic rational frameworks, their construction relies heavily on labor-intensive domain expertise, creating an ``expert bottleneck" that hinders scalability in general scenarios. To bridge the gap between the generalization capabilities of LLMs and the rigor of decision theory, we propose Doc2AHP, a novel structured inference framework guided by AHP principles. Eliminating the need for extensive annotated data or manual intervention, our approach leverages the structural principles of AHP as constraints to direct the LLM in a constrained search within the unstructured document space, thereby enforcing the logical entailment between parent and child nodes. Furthermore, we introduce a multi-agent weighting mechanism coupled with an adaptive consistency optimization strategy to ensure the numerical consistency of weight allocation. Empirical results demonstrate that Doc2AHP not only empowers non-expert users to construct high-quality decision models from scratch but also significantly outperforms direct generative baselines in both logical completeness and downstream task accuracy.
\end{abstract}

\section{Introduction}

Multi-Criteria Decision Making (MCDM) is pivotal for addressing complex real-world challenges \cite{mardani2015multiple}. Within this framework, the Analytic Hierarchy Process (AHP) \cite{saaty1977,saaty2008decision} has established itself as a prevalent methodology in fields ranging from infrastructure development \cite{vaidya2006analytic} and medical assessment \cite{liberatore2008medical} to cybersecurity \cite{Moreira2025Cybersecurity}, owing to its capability to decompose abstract objectives into hierarchical criteria and quantify weights via pairwise comparisons. However, constructing high-quality AHP models traditionally relies on domain experts to manually synthesize extensive documentation to establish these hierarchies—a process that is inherently labor-intensive and difficult to scale.

The ascent of Large Language Models (LLMs) has rendered the automated construction of decision models feasible, capitalizing on their extensive knowledge bases. Early ``virtual expert" approaches \cite{kampourakis2025llmassistedahpexplainablecyber,svoboda2024enhancingmulticriteriadecisionanalysis,lu2024ahp} relying solely on parametric knowledge allow for direct generation but suffer from critical limitations. First, being detached from concrete evidence (e.g., industry standards or technical reports), the generated criteria often remain generic and superficial. Second, in quantitative tasks, LLMs are prone to hallucinations and output instability \cite{ji2023survey}. The opacity of such black-box generation precludes evidentiary support, rendering it inadequate for high-stakes decision-making where traceability is imperative \cite{rudin2019stop}.

To address these challenges, we propose Doc2AHP, a novel automated modeling framework designed to reason structurally rigorous and interpretable AHP models directly from unstructured document collections. Our core insight is that the geometric structure within the semantic space often latently encodes the hierarchical relationships of decision criteria. Consequently, departing from previous methods that rely heavily on direct Prompt Engineering, we introduce specific structural priors to guide the modeling process. Specifically, Doc2AHP operates in two phases: (1) Structure Generation, which mines the semantic geometry via hierarchical clustering to instantiate the AHP skeleton directly from the document distribution; and (2) Weight Estimation, which employs a Leader-Guided Multi-Agent Collaborative Mechanism. By orchestrating diverse agents for independent sampling and collective debate, this mechanism effectively mitigates stochastic errors and enforces mathematical consistency constraints.

The main contributions of this work are as follows: \begin{itemize} \item We propose Doc2AHP, an end-to-end framework leveraging semantic geometry to guide the automated transformation of unstructured documents into computable decision models. \item We introduce a multi-agent collaborative strategy with adaptive consistency optimization, effectively resolving the quantitative reasoning inconsistencies inherent to LLMs. \item Empirical results on IMDb, HotelRec, and Beer Advocate demonstrate that Doc2AHP significantly outperforms direct generation baselines in both structural rationality and downstream decision accuracy. \end{itemize}

\section{Related Work}

\subsection{Large Language Models for Complex Reasoning and Decision Support}

LLMs have demonstrated remarkable potential in complex reasoning, exemplified by Chain-of-Thought (CoT) \cite{wei2022chain} and Tree of Thoughts (ToT) \cite{yao2023tree}, which enhance multi-step logic through intermediate reasoning and tree search. However, professional decision support requires greater structural integrity than open-domain tasks. To address this, recent research employs \textit{structural inductive bias} by either extracting explicit structures—such as knowledge graphs in GraphRAG \cite{edge2024local}—or constraining generation with classical models like decision trees \cite{ye2025llm} and Finite State Machines \cite{canoneurosymbolic}.

Complementing these structural approaches, Multi-Agent and Human-AI collaboration have emerged to enhance robustness. Key innovations include cascaded frameworks for dynamic task allocation \cite{fanconi2025cascaded}, multi-agent debate mechanisms to rectify cognitive biases \cite{choi2025debatevoteyieldsbetter}, and confidence calibration in RAG systems \cite{jang2025reliable}. Motivated by these trends of \textit{structural synthesis} and \textit{collaborative augmentation}, Doc2AHP extends this paradigm to Multi-Criteria Decision Making (MCDM). Unlike prior generic reasoning tasks, we leverage the hierarchical topology of AHP as a rigorous constraint combined with multi-agent estimation to achieve end-to-end, interpretable model construction from unstructured documents.

\subsection{Automated Construction of AHP Models}

The Analytic Hierarchy Process (AHP) is a standard for structured decision-making but faces a scalability bottleneck due to its reliance on manual expert modeling. To address this, recent research utilizes LLMs as ``Virtual Experts.'' For instance, Kampourakis et al. \cite{kampourakis2025llmassistedahpexplainablecyber} and Svoboda et al. \cite{svoboda2024enhancingmulticriteriadecisionanalysis} simulate expert panels to automate pairwise comparisons for cybersecurity. Similarly, Lu et al. \cite{lu2024ahp} apply AHP to evaluate Open-ended QA, while Wang et al. \cite{wang2025allgeneralframeworkllmsbased} enhance domain specialization via LoRA fine-tuning. Bang et al. \cite{bang2025transforminguserdefinedcriteria} further combine LLM scoring with statistical metrics for interpretable recommendations.

However, these approaches predominantly rely on direct prompting or expensive Supervised Fine-Tuning (SFT). This ``end-to-end'' generation often neglects intrinsic logical constraints, making hierarchies prone to hallucinations and pairwise matrices inconsistent. In contrast, Doc2AHP introduces hierarchical clustering as a structural prior, leveraging document geometry to ensure logical integrity in an unsupervised manner. Furthermore, we replace the single-expert perspective with Multi-Agent Collaboration, employing consensus optimization to ensure mathematical rigor and robustness under zero-shot conditions.

\section{Methodology}

\subsection{Overview of the Proposed Framework}

In this paper, we propose Doc2AHP. Formally, given a decision objective $c_0$, a collection of relevant documents $D$, and a set of alternatives $A$, the framework derives a viable decision outcome. Our methodology comprises two tightly coupled phases: Probabilistic AHP Construction and Decision Inference.

\begin{figure*}
    \centering
    \includegraphics[width=1\linewidth]{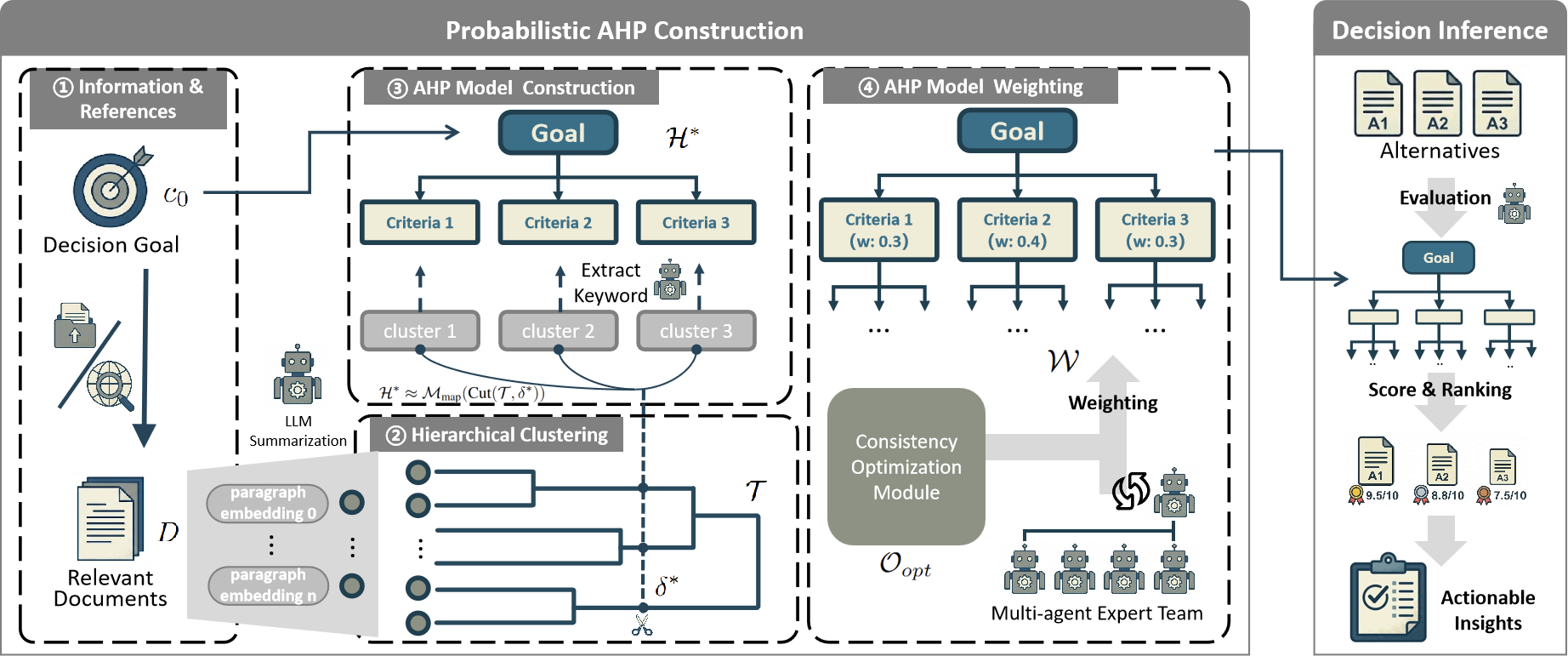}
    \caption{The pipeline of the Doc2AHP.}
    \label{fig:1}
\end{figure*}

\subsection{Probabilistic AHP Construction}

The objective of this phase is to estimate the AHP model parameters—the hierarchical structure $\mathcal{H}$ and weight matrix $\mathcal{W}$—conditioned on context $\{c_0, D\}$. We formulate this as a joint probability estimation problem, decomposed into structure generation and weight estimation:
\begin{equation}
p(\mathcal{H}, \mathcal{W} | c_0, D ) \propto p_{\theta}(\mathcal{W} | \mathcal{H}, c_0, D) \cdot p_{\theta}(\mathcal{H} | c_0, D).
\end{equation}

Theoretically, the optimal structure $\mathcal{H}^*$ maximizes the posterior probability over the entire tree space. However, exact inference is NP-Hard due to the combinatorial explosion of the search space with respect to document fragments. To address this, we employ a heuristic strategy that utilizes a hierarchical clustering-based skeleton as a structural prior to constrain the search space.

We posit that semantically correlated documents should be consolidated under the same sub-criterion\cite{sanderson1999deriving} within the decision hierarchy. Consequently, we model structure generation as a deterministic mapping from the document hierarchical clustering tree $\mathcal{T} = \text{Cluster}(D)$ to the AHP decision tree $\mathcal{H}$. The optimal structure $\mathcal{H}^*$ is approximated by identifying a set of cut points on $\mathcal{T}$ that maximize both semantic coverage and independence:
\begin{equation}
\mathcal{H}^* \approx \mathcal{M}_{\text{map}}(\text{Cut}(\mathcal{T}, \delta^*)),
\end{equation}
where $\mathcal{M}_{\text{map}}$ denotes the mapping function that transforms cluster nodes into AHP criteria. This formulation transforms the intricate structure generation problem into a problem of recursive traversal and pruning on the clustering tree.

Building upon the determined structure $\mathcal{H}^*$, and to address the instability and logical hallucinations inherent in direct numerical outputs from LLMs, we formulate the weight estimation phase as an ``Aggregation-Optimization'' process driven by multi-agent collaboration. This mechanism is designed to neutralize individual biases by leveraging collective intelligence, while enhancing the mathematical rigor of the final outcomes via constrained optimization.

Initially, a panel of agents $\{Agent_k\}_{k=1}^K$ independently constructs a set of pairwise comparison matrices $\{M^{(k)}\}$ based on the document evidence $D$ and the established structure $\mathcal{H}^*$, thereby capturing semantic judgments from diverse perspectives. Subsequently, we employ an aggregation operator (e.g., the geometric mean) \cite{ernest1998aggregating} to synthesize these multi-agent judgments into a unified consensus matrix $\bar{M}$, effectively mitigating stochastic noise.

However, the consensus matrix $\bar{M}$ may still fail to satisfy the transitivity axiom (i.e., Consistency Ratio $CR > 0.1$\cite{saaty1987analytic}). To address this, we introduce an optimization mapping $\mathcal{O}_{opt}$. This mapping seeks a rectified matrix $M^*$ within the consistency space $\mathcal{M}_{consistent}$ that is closest to $\bar{M}$, thereby enforcing logical constraints while preserving the semantic consensus of the group:
\begin{equation}
M^* = \mathcal{O}_{opt}(\bar{M}) = \mathop{\arg\min}_{M \in \mathcal{M}_{consistent}} \mathcal{D}(M, \bar{M}).
\end{equation}
The final weight vector $\mathbf{w}^*$ is derived from the principal eigenvector of the rectified matrix $M^*$.

\subsection{Decision Inference}

Having established the optimal AHP structure $\mathcal{H}$ and the weight vector $\mathbf{w}$, we advance the framework to the Decision Inference phase. The primary objective of this stage is to quantify the set of alternatives $\mathcal{A} = \{a_1, \dots, a_M\}$ and generate an interpretable decision report.

Specifically, for any arbitrary alternative $a_k \in \mathcal{A}$ and leaf criterion $c_j \in \mathcal{C}_{leaf}$, the task is to estimate the local utility score $s_{kj}$. Unlike traditional AHP which relies on manual elicitation, we employ an LLM as an automated evaluator. We formulate $s_{kj}$ as the conditional expectation given the criterion definition, the alternative description $d(a_k)$, and the relevant document evidence $D$:
\begin{equation}
s_{kj} = \mathbb{E}_{y \sim p_\theta(\cdot | c_j, a_k, D)} \left[ \mathcal{M}_{\text{score}}(y) \right],
\end{equation}
where $p_\theta$ represents the pre-trained LLM, which generates a natural language output $y$ encompassing both the evaluation rationale and a quantitative rating. Furthermore, $\mathcal{M}_{\text{score}}$ denotes a deterministic parsing function that maps this natural language output onto a standardized AHP numerical scale.

Given the scoring matrix $\mathbf{S}$ and weight vector $\mathbf{w}$, the composite utility score $U(a_k)$ is computed via standard AHP aggregation. This yields the final decision result set $\mathcal{R} = \{(a_k, U(a_k))\}_{k=1}^M$, enabling the ranking of alternatives based on their utilities.

To facilitate downstream decision-making tasks, such as comparative analysis and report drafting, we leverage the quantitative findings for data-to-text generation. Formally, the decision report $Y_{\text{report}}$ is modeled as a sample drawn from a distribution conditioned on the document evidence $D$, the AHP structure $\mathcal{H}$, and the evaluation results $\mathcal{R}$:

\begin{equation}
Y_{report} \sim p_\theta(Y \mid \mathcal{R}, \mathcal{H}, D).
\end{equation}

The pipeline of Doc2AHP is shown in Fig. \ref{fig:1}. The LLM prompt templates employed in the process can be found in Appendix B. The subsequent subsections will elaborate on each of these steps. 

\subsection{Document Embedding and Semantic Representation}

The process initiates by mapping the unstructured document collection $\mathcal{D}$ into a high-dimensional vector space.  Specifically, each document $d_i \in \mathcal{D}$ is partitioned into fine-grained semantic units, denoted as $p_{i,j}$, which are then encoded into $k$-dimensional vectors $\mathbf{v}_{i,j} \in \mathbb{R}^k$ via a pre-trained embedding model. These vectors constitute the semantic space $\mathcal{V}$, serving as the foundational basis for the subsequent hierarchical clustering and structure generation.

\subsection{Semantic Hierarchical Clustering}

To explicitly capture the geometric structure inherent in the document information, we perform hierarchical clustering on the embedding collection $\mathcal{V}$.

We adopted Ward's method \cite{ward1963hierarchical} to minimize intra-cluster variance, ensuring that the resulting clusters exhibit high semantic compactness. This property is particularly critical for constructing coherent and logically consistent AHP criteria.

Formally, this process induces a binary tree $\mathcal{T} = (\mathcal{N}, \mathcal{E})$, where leaf nodes represent primitive paragraph units, $\mathcal{E}$ represents the parent-child relationship and each internal node $u \in \mathcal{N}$ represents the set of paragraphs $\mathcal{P}_u$ subsumed within its subtree. The height $h(u)$ of a node $u$ quantifies the semantic distance underlying the specific merger at that level.

We adopt paragraphs, rather than sentences, as the fundamental unit of analysis. Acting as natural semantic units that encapsulate self-contained concepts, paragraphs circumvent the semantic fragmentation inherent in sentence-level clustering. This choice aligns more effectively with the level of abstraction required for decision criteria. The resulting tree $\mathcal{T}$ serves directly as the structural skeleton for the subsequent structure generation phase.

\subsection{Recursive AHP Structure Generation}

Leveraging the geometric skeleton provided by the semantic clustering tree $\mathcal{T}$, this phase instantiates the AHP hierarchy $\mathcal{H}$ via a top-down recursive process. We define a mapping function $\Phi: \mathcal{C} \rightarrow \mathcal{N}$ to align each generated AHP criterion $c$ with its corresponding semantic cluster $u$ within $\mathcal{T}$. This alignment ensures that the structural logic of the hierarchical criteria remains grounded in the underlying semantic distribution of the document collection.

To ensure that the generated decision model adheres to AHP methodological norms (e.g., a clear hierarchy and concise indicators) while adapting to the varying complexity demands of specific tasks, we introduce explicit complexity constraints to govern the pruning and mapping processes. In contrast to standard clustering approaches that rely solely on a fixed distance threshold $\delta$, Doc2AHP employs the maximum branching factor ($K_{max}$) and the maximum hierarchical depth ($D_{max}$) as core hyperparameters. These parameters regulate the width and depth of the decision hierarchy, thereby preventing cognitive overload.

The system operates under two parameter configuration modes: (1) Expert-Specified Mode: Users directly prescribe $K_{max}$ (typically within the range of $5$--$9$) and $D_{max}$ based on domain expertise. (2) Adaptive Recommendation Mode: For non-expert users, the system leverages an LLM to analyze the complexity of the decision objective $c_0$, automatically inferring the optimal $K_{max}$ and $D_{max}$.

The specific generation process proceeds as follows:
\begin{itemize}
    \item For a given AHP node $c$ and its corresponding cluster $u = \Phi(c)$, provided that the current depth $d < D_{max}$, we execute a split operation on the clustering subtree rooted at $u$. The objective is to identify a set of sub-clusters $\mathcal{S} = \{u_1, \dots, u_m\}$ such that the number of child nodes satisfies $2 \le m \le K_{max}$, while maximizing the separation of these sub-clusters within the semantic space. Essentially, this involves dynamically locating the optimal natural semantic breakpoints within the local subtree, subject to the defined constraints.

    \item Once the set of sub-clusters $\{u_i\}_{i=1}^m$ is determined, we transform these geometric clusters into symbolic AHP criteria:
    
    \textbf{1) Criterion Summarization:} Given that $u_i$ aggregates a set of paragraph vectors, we employ an LLM to summarize the textual content covered by these vectors, generating a concise criterion label $label_i$:
    \begin{equation}
        c_i \leftarrow \text{LLM}_{\text{gen}}(\text{Text}(u_i) \mid \text{Context}=\mathcal{P}_{u_i}).
    \end{equation}
    
    \textbf{2) Semantic Verification:} To prevent semantic drift, we introduce a discriminator $\text{LLM}_{\text{verify}}$ to calculate the relevance score between the generated child criterion $c_i$ and its parent $c$. If the score falls below a semantic threshold $\tau$, the branch is either pruned or re-merged.
\end{itemize}

The recursion iterates until the maximum depth $D_{max}$ is reached or further effective splitting becomes infeasible (i.e., $m < 2$). The detailed pseudo-code is presented in Algorithm \ref{alg:ahp_hierarchy_construction}.

\begin{algorithm}[tb]
\caption{Structure Generation with Cognitive Constraints}
\label{alg:ahp_hierarchy_construction}
\textbf{Input}: Cluster Tree $\mathcal{T}$, Decision Goal $c_0$, Documents $\mathcal{D}$\\
\textbf{Hyperparameters}: Max branching $K_{max}$, Max depth $D_{max}$, Relevance threshold $\tau$\\
\textbf{Output}: AHP Hierarchy $\mathcal{H}$

\begin{algorithmic}[1]
\STATE // Initialization: Set constraints via User or LLM
\IF{parameters not provided}
    \STATE $K_{max}, D_{max} \leftarrow \text{LLM}_{\text{infer\_complexity}}(c_0, \mathcal{D})$
\ENDIF

\STATE Initialize $\mathcal{H}$ with root $c_0$, $\Phi(c_0) \leftarrow \text{root}(\mathcal{T})$
\STATE \textbf{Queue} $Q \leftarrow \{(c_0, \text{depth}=0)\}$

\WHILE{$Q$ is not empty}
    \STATE $(c_{curr}, d) \leftarrow Q.\text{pop}()$
    \STATE $u_{curr} \leftarrow \Phi(c_{curr})$
    
    \STATE // Stop condition: Depth limit or Leaf node
    \IF{$d \ge D_{max}$ \OR $u_{curr}$ is leaf}
        \STATE \textbf{continue}
    \ENDIF

    \STATE // Find best $m$ clusters subject to $m \le K_{max}$
    \STATE $\{u_1, \dots, u_m\} \leftarrow \text{AdaptiveCut}(u_{curr}, \mathcal{T}, K_{max})$
    
    \FOR{$i = 1$ to $m$}
        \STATE // Abstract semantics to text
        \STATE $txt \leftarrow \text{RetrieveTexts}(u_i, \mathcal{D})$
        \STATE $name_i \leftarrow \text{LLM}_{\text{gen}}(txt, \text{parent}=c_{curr})$
        
        \STATE // Semantic Verification
        \IF{$\text{LLM}_{\text{verify}}(name_i, c_{curr}) \ge \tau$}
            \STATE Create node $c_i$ with label $name_i$
            \STATE Add edge $(c_{curr}, c_i)$ to $\mathcal{H}$
            \STATE $\Phi(c_i) \leftarrow u_i$
            \STATE $Q.\text{push}((c_i, d+1))$
        \ENDIF
    \ENDFOR
\ENDWHILE

\STATE \textbf{return} $\mathcal{H}$
\end{algorithmic}
\end{algorithm}

\subsection{Consistency-Aware Weight Estimation}

With the structure $\mathcal{H}$ established, the process transitions to the weight estimation phase. To mitigate the instability inherent in direct numerical outputs from LLMs, we propose a Leader-Guided Collaborative Optimization mechanism. This mechanism is designed to distill objective weights from multi-perspective subjective judgments, ensuring mathematical consistency while maintaining alignment with expert intent.

To counteract single-perspective random errors and incorporate high-level strategic guidance, we devise a hierarchical agent architecture comprising $K$ Domain Expert Agents and a singular Leader Agent.
Initially, the panel of Expert Agents independently constructs pairwise comparison matrices, denoted as $M^{(k)} \in \mathbb{R}^{n \times n}$, drawing upon the document evidence $D$ and the established hierarchy $\mathcal{H}$. To synthesize collective intelligence, we employ the Weighted Geometric Mean as the aggregation operator:
\begin{equation}
    \bar{a}_{ij} = \prod_{k=1}^K \left( a_{ij}^{(k)} \right)^{\gamma_k}, \quad \text{s.t.} \sum \gamma_k = 1,
\end{equation}
where $\bar{M} = [\bar{a}_{ij}]$ represents the aggregated consensus matrix, and $\gamma_k$ denotes the confidence weight assigned to the $k$-th agent (defaulting to $1/K$). Although $\bar{M}$ captures the central tendency of the group, it frequently fails to satisfy consistency requirements (specifically transitivity, i.e., $CR > 0.1$) due to inherent noise.

To address the aforementioned inconsistency, we formulate weight estimation as a constrained optimization problem. While classical algebraic \cite{cao2008modifying} and heuristic \cite{sarani2024optimizing,girsang2015repairing} approaches effectively rectify inconsistent matrices, they often prioritize numerical convergence ($CR<0.1$) over the preservation of original preference information, risking the distortion of expert intent \cite{saaty2002hard,andrecut2014decision}. In the context of LLM-based multi-agent systems, where judgments exhibit both inherent stochasticity and high-level reasoning, such blind numerical adjustments are inadequate as they ignore the semantic validity of the decision. To address this, we propose a Leader-Guided Collaborative Optimization framework rooted in the Logarithmic Least Squares Method (LLSM) \cite{kou2017intelligent}. By integrating the Leader Agent’s strategic directives as explicit constraints within a convex optimization model, our approach ensures that the derived weights are not only statistically robust against agent hallucinations but also semantically aligned with domain expertise, effectively bridging the gap between mathematical rigor and logical coherence.

The goal is to identify an optimal weight vector $\mathbf{w}^*$ that minimizes the discrepancy between its induced consistent matrix and the consensus matrix $\bar{M}$, while satisfying the semantic constraints imposed by the Leader Agent.

We employ the Logarithmic Least Squares Method (LLSM) framework, which defines the optimization objective as minimizing the Euclidean distance within the logarithmic space. To integrate high-level guidance, natural language recommendations from the Leader Agent (e.g., ``Safety should be significantly more important than Cost'') are translated into a set of linear inequality constraints, denoted as $\Omega_{leader}$. The optimization model is formulated as follows:
\begin{equation}
\begin{aligned}
\mathbf{w}^* = 
\mathop{\arg\min}_{\mathbf{w}} \quad &
\sum_{i=1}^n \sum_{j=1}^n 
\left( \ln \bar{a}_{ij} - \ln \frac{w_i}{w_j} \right)^2 \\
\text{s.t.} \quad
& \sum_{i=1}^n w_i = 1, \quad w_i > 0, \\
& w_i \ge \beta_{ij} w_j, \quad \forall (i,j,\beta) \in \Omega_{leader}. 
\end{aligned}
\end{equation}

 Specifically, $\Omega_{leader}$ denotes the set of hard constraints derived by the Leader Agent based on domain knowledge (e.g., $\beta_{ij}=3$ implies that criterion $i$ is at least three times as important as $j$). This formulation constitutes a typical convex optimization problem, guaranteeing the existence of a global optimum. Solving this problem yields a set of robust weights that statistically align with the group consensus while logically adhering to the Leader's guidance. Finally, to preserve the underlying semantics of the comparative magnitudes, the optimized ratios are mapped onto a discrete space compliant with the standard 1--9 AHP scale.

\subsection{Decision Inference and Interpretability Support}

Once the optimized AHP model $\mathcal{M}^* = (\mathcal{H}, \mathbf{w}^*)$ is constructed, the process proceeds to the final stage of decision inference. During this phase, the LLM is employed to map unstructured alternative descriptions into computable utility scores, thereby providing a structured foundation for subsequent interpretability analysis.

For each alternative $a_k \in \mathcal{A}$ and leaf criterion $c_j$, the LLM evaluates a local score $s_{kj} \in [0, 1]$ grounded in criterion definitions and document evidence. The final aggregated score $U(a_k)$ is then computed via the linear aggregation of criterion weights, which determines the ranking of alternatives to provide decision-makers with mathematically rigorous priority recommendations.

Beyond numerical rankings, a salient advantage of Doc2AHP lies in its intrinsic interpretability. Unlike end-to-end black-box recommenders, our framework yields a complete decision chain, spanning from raw document evidence to the hierarchical criteria $\mathcal{H}$ and ultimately to the quantified weights $\mathbf{w}^*$. Given such explicit structures and scored alternatives, producing human-readable decision reports is a relatively straightforward downstream step rather than a central modeling challenge. Therefore, while the initial problem formulation (Section 3.3) includes report generation for the sake of task completeness, this study maintains a focused emphasis on the primary scientific challenges: structure induction and consistent weighting.

\section{Experiments}

\subsection{Experimental Setup}

\noindent\textbf{Datasets \& Benchmark.} Unlike large-scale retrieval, AHP targets high-stakes decision problems within a restricted consideration set. Therefore, we constructed \textbf{DecisionBench}, which comprises six decision scenarios across the movie, hotel, and beer domains (see Table \ref{tab:decisionbench}). To simulate realistic, fine-grained decision-making, we implemented LLM-driven profile augmentation to enrich the descriptive details based on the \textbf{IMDb} \cite{maas2011learning}, \textbf{HotelRec} \cite{antognini-faltings:2020:LREC1}, and \textbf{Beer Advocate} \cite{mcauley2013amateurs} datasets (augmentation details provided in the Appendix A2). For each scenario, we created a stratified shortlist of 20 candidates ($N=20$) by sampling across distinct ground truth quality levels (sampling details provided in Appendix A1). This design focuses the evaluation on the model's capability for nuanced value trade-offs under multiple criteria.

\noindent\textbf{Baselines.} We compare \textbf{Doc2AHP} against two representative paradigms: (1) \textbf{Standard-AHP} \cite{lu2024ahp}: A single-LLM CoT reasoning framework. Since its original ranking heuristic inherently bypasses consistency checks, we adapted it to direct matrix generation to enable comparable numerical evaluation. This baseline serves as a proxy for unconstrained single-model reasoning. (2) \textbf{Debate-AHP} \cite{svoboda2024enhancingmulticriteriadecisionanalysis}: A multi-agent collaboration framework where diverse roles iteratively refine criteria. We aligned the input setting by injecting the same domain documents.  While both methods utilize multi-agent interactions, Debate-AHP relies on unstructured semantic negotiation (e.g., voting) to reach consensus. In contrast, Doc2AHP incorporates structural priors for criteria construction and enforces mathematical constraints (e.g., consistency optimization) to guide the multi-agent weighting process.

\noindent\textbf{Implementation Details.} To decouple retrieval variance, we adopted a ``Fixed Domain Context'' protocol. All methods receive the same standardized corpus of approximately 20 expert documents as input (details in the Appendix A3). To comply with context window limits, we uniformly preprocessed all input documents (details in the Appendix A4). Unless otherwise specified, all experiments use GPT-5.2 (Temp=0.1) as the core reasoning model. we also conducted robustness tests on the Llama-3.1 model series (8B and 70B). Each experiment involved an independent execution of the full pipeline.

\noindent\textbf{Evaluation Metrics.}
We employ NDCG@5 as the primary metric to measure the quality of top-tier recommendations, and NDCG@10 to evaluate the robustness of the overall ranking \cite{jarvelin2000ir}. Regarding numerical reliability, we report CR-max, CR-mean, and the Pass Rate to evaluate the consistency of pairwise comparison matrices across runs, where the pass rate is defined as the proportion of matrices satisfying the threshold $CR < 0.1$.

\begin{table}[t]
\centering
\caption{Overview of DecisionBench Scenarios}
\label{tab:decisionbench}
\small
\setlength{\tabcolsep}{5pt}
\renewcommand{\arraystretch}{1.05}
\begin{tabular}{l l p{3.4cm} p{1.5cm}}
\hline
\textbf{Domain} & \textbf{Scen. ID} & \textbf{Decision Goal} & \textbf{Key Focus} \\
\hline
Movies & M-Act  & ``Select Best Visual \& Technical Blockbuster''        & Visuals, Tech \\
       & M-Dra  & ``Select Best Narrative \& Artistic Drama''            & Narrative, Depth \\
Hotels & H-Fam  & ``Best Family-Oriented Hotel''                         & Safety, Amenities \\
       & H-Bus  & ``Best Business \& Functional Hotel''                  & Efficiency, Loc \\
Beers  & B-Ref  & ``Best Sessionable Summer Beer''                       & Crispness \\
       & B-Cplx & ``Evaluate Beer for Complexity, Body \& Aroma''         & Flavor Depth \\
\hline
\end{tabular}
\end{table}

\subsection{End-to-End Decision Accuracy and Domain Alignment}

\begin{table*}[t]
\centering
\caption{Performance comparison on three datasets with two scenarios using NDCG@5 and NDCG@10.}
\label{tab:ndcg_results}
\resizebox{\textwidth}{!}{
\begin{tabular}{l cc cc cc cc cc cc}
\hline
\multirow{3}{*}{Method}
& \multicolumn{4}{c}{IMDb} 
& \multicolumn{4}{c}{HotelRec} 
& \multicolumn{4}{c}{Beer-Advocate} \\
\cline{2-13}

& \multicolumn{2}{c}{M-Act} 
& \multicolumn{2}{c}{M-Dra}
& \multicolumn{2}{c}{H-Fam} 
& \multicolumn{2}{c}{H-Bus}
& \multicolumn{2}{c}{B-Ref} 
& \multicolumn{2}{c}{B-Cplx} \\
\cline{2-13}

& @5 & @10 & @5 & @10 
& @5 & @10 & @5 & @10 
& @5 & @10 & @5 & @10 \\
\hline

Standard-AHP    & 0.817 & 0.841 & 0.830 & 0.844 & 0.950 & 0.964 & 0.968 & \textbf{0.982} & 0.948 & 0.962 & 0.948 & 0.960 \\

Debate-AHP      & 0.778 & \textbf{0.845} & 0.777 & 0.831 & 0.912 & 0.900 & \textbf{0.978} & 0.977 & 0.942 & 0.958 & 0.949 & 0.956 \\

Ours            & \textbf{0.839} & 0.839 & \textbf{0.854} & \textbf{0.871} & \textbf{0.974} & \textbf{0.965} & 0.975 & \textbf{0.982} & \textbf{0.964} & \textbf{0.966} & \textbf{0.965} & \textbf{0.969} \\

\hline
\end{tabular}}
\end{table*}

This experiment aims to verify whether \textbf{Doc2AHP} can capture fine-grained quality features according to specific decision goals. Table \ref{tab:ndcg_results} presents the full experimental results. Overall, \textbf{Doc2AHP} achieved optimal performance on the NDCG@5 metric across the vast majority of scenarios (5/6).

The data demonstrates \textbf{Doc2AHP}'s significant advantage in NDCG@5. For instance, in M-Dra (Drama), our method reached 0.854, significantly outperforming \textbf{Standard-AHP} (0.830) and \textbf{Debate-AHP} (0.777). This proves that introducing structured constraints effectively filters out decision noise. In contrast, \textbf{Debate-AHP} showed weak performance on the M-Act @5 metric (0.778), indicating that multi-agent negotiation is prone to a ``compromise effect," making it difficult to accurately identify top-tier candidates with unique styles.

A comparative analysis reveals that as task semantic complexity increases, \textbf{Doc2AHP}'s advantage becomes more pronounced. In H-Bus (Business Hotel), where standards are explicit, the performance gap between methods is minimal; however, in scenarios involving abstract concepts like M-Dra (Narrative Depth) and B-Cplx (Beer Complexity), \textbf{Doc2AHP} exhibits dominant performance. Lacking hierarchical semantic extraction, \textbf{Standard-AHP} tends to lose these deep evaluation dimensions when processing long documents, whereas our semantic clustering module successfully achieves precise alignment in complex domains.

Notably, on the M-Act NDCG@10 metric, \textbf{Doc2AHP} (0.839) scored slightly lower than \textbf{Debate-AHP} (0.845). This suggests that baseline methods tend to fit ``mass popularity,'' thereby easily scoring well in broad rankings. However, \textbf{Doc2AHP} strictly adheres to professional standards within the documents, potentially assigning lower scores to movies that are ``high-grossing but technically mediocre.'' In AHP scenarios, this strict ``quality over quantity'' approach ensures the professional caliber of the Top-5 recommendations.

\subsection{Mathematical Validity of Numerical Reasoning and Cross-Model Robustness}

\begin{table*}[t]
\centering
\caption{Consistency comparison under different LLM generators.}
\label{tab:consistency_results}
\setlength{\tabcolsep}{4pt}
\begin{tabular}{lccc ccc ccc}
\toprule
Method 
& \multicolumn{3}{c}{Llama3.1-8b} 
& \multicolumn{3}{c}{Llama3.1-70b} 
& \multicolumn{3}{c}{GPT-5.2} \\
\cmidrule(lr){2-4} \cmidrule(lr){5-7} \cmidrule(lr){8-10}

& CR mean & CR max & Pass rate 
& CR mean & CR max & Pass rate 
& CR mean & CR max & Pass rate \\
\midrule

Standard-AHP 
& 0.1693 & 0.7514 & 71.42\% 
& 0.1414 & 1.3921 & 86.95\% 
& 0.0434 & 0.3167 & 88.57\% \\

Debate-AHP   
& 0.2902 & 0.4309 & 0.00\%  
& \textbf{0.0065} & \textbf{0.0132} & \textbf{100\%}   
& \textbf{0.0114} & \textbf{0.0335} & \textbf{100\%}   \\

Ours         
& \textbf{0.0139} & \textbf{0.0807} & \textbf{100\%}   
& 0.0530 & 0.0957 & \textbf{100\%}   
& 0.0326 & 0.0882 & \textbf{100\%}   \\
\bottomrule

\end{tabular}
\end{table*}

To evaluate numerical robustness, we conduct a stress test using H-Fam scenario documents as input and test three models with increasing parameter scales (Llama-3.1-8B,  Llama-3.1-70B and GPT-5.2 ). Each baseline is run five times independently. Table \ref{tab:consistency_results} presents the performance of each method across different model scales. The results reveal the decisive advantage of Doc2AHP in ensuring mathematical rigor.

In the weak model environment of Llama-3.1-8B, \textbf{Doc2AHP} maintained a 100\% pass rate (Mean CR=0.0139), whereas baseline methods exhibited significant performance degradation. The most striking observation is the performance cliff of \textbf{Debate-AHP}, which dropped from a 100\% pass rate on the 70B model to 0\% on the 8B model. This phenomenon indicates that the multi-agent debate mechanism relies heavily on the instruction-following and logical reasoning capabilities of the foundation model. With weak models, the ``negotiation" between agents fails, leading to ``Consensus Collapse"—where multiple hallucinating agents mislead one another. Instead of correcting errors, this amplifies numerical inconsistency (Mean CR surged to 0.2902). In contrast, \textbf{Doc2AHP} does not rely on the model to ``spontaneously" generate perfect values; instead, it employs a consistency optimization algorithm as a final ``gatekeeper." Regardless of the base model's weakness, as long as it outputs baseline values, the optimization module forces convergence into a mathematically valid range.

\textbf{Standard-AHP} exhibited fluctuating pass rates (71\%-87\%) on both Llama models, with the Mean CR consistently failing to meet the acceptance standard (exceeding 0.14). Notably, it demonstrated extreme instability: on the Llama-3.1-70B model, the Max CR reached as high as 1.3921. This implies that without external feedback loops, a single LLM is prone to generating decision matrices with chaotic logic. This further validates the necessity of introducing computational constraint modules.

\subsection{Ablation Study}

To isolate the contribution of each component, we compare the full model against two variants on the H-Fam scenario: (1) \textbf{w/o Struct}, which replaces the hierarchical clustering with standard linear summarization (identical to baselines) to test the necessity of structural priors; and (2) \textbf{w/o Opt}, which bypasses the consistency check loop to generate pairwise matrices in a single pass, evaluating the impact of algorithmic constraints.

\begin{table}[t]
\centering
\caption{Ablation results of different model variants.}
\label{tab:ablation}
\begin{tabular}{lcc}
\hline
Model Variant & NDCG@5 & NDCG@10 \\
\hline
w/o Struct & 0.986 & \textbf{0.963} \\
w/o Opt & 0.977 & 0.953 \\
\hline
Full Model (Ours) & \textbf{0.993} & 0.961 \\
\hline
\end{tabular}
\end{table}

Table \ref{tab:ablation} presents the ablation results. Comparing the Full Model with \textbf{w/o Struct}, we observe a notable phenomenon: the Full Model achieved high precision on NDCG@5 (0.993), significantly outperforming the variant (0.986); conversely, on NDCG@10, \textbf{w/o Struct} held a slight lead (0.963 vs. 0.961). This suggests that removing semantic clustering causes the model to extract more ``generic" criteria. While beneficial for covering a broader candidate set (@10), this approach lacks sufficient sensitivity and discriminative power when identifying top-tier candidates (@5). The structural module ensures decision ``stringency" via fine-grained criteria extraction, thereby achieving superior top-tier recommendations.

The \textbf{w/o Opt }variant performed worst on both metrics (0.977 / 0.953). This confirms that without Consistency Checks, matrices generated directly by LLMs are prone to logical conflicts, leading to weight distortion. The consistency optimization module effectively acts as a mathematical gatekeeper, preventing numerical hallucinations from compromising the final ranking.

\section{Conclusion}

We presented Doc2AHP, a framework that bridges unstructured LLM reasoning with rigorous multi-criteria decision analysis. By enforcing algorithmic constraints, we effectively decoupled complex decision logic from the model's parametric knowledge. Our evaluations on DecisionBench confirm that this neuro-symbolic approach not only achieves superior accuracy in identifying top-tier candidates (NDCG@5) but also ensures mathematical robustness even on lightweight models (e.g., Llama-8B). This establishes a promising path for deploying auditable, expert-level decision systems on compute-constrained edge devices.

While building explicit hierarchical structures is more time-consuming than direct prompting, this cost is justified in high-stakes domains where reliability outweighs speed. Doc2AHP shifts decision-making from intuitive fast thinking to logically verifiable slow thinking, enhancing the robustness of automated decision processes.

\appendix

\bibliographystyle{named}
\bibliography{ijcai26}

\end{document}